\documentclass[journal, onecolumn, draftcls, 10pt]{IEEEtran}
\usepackage{graphicx} 
\usepackage{amsmath,amssymb, amsfonts}
\usepackage{cite} 
\usepackage{float}
\usepackage{color}
\usepackage{algorithm}
\usepackage{algorithmic, bm, textcomp}
\usepackage{hyperref}

\def \va{\mathbf {a}}
\def \vc{\mathbf {c}}
\def \vf{\mathbf {f}}
\def \vx{\mathbf {x}}

\def \vv{\mathbf {v}}
\def \vw{\mathbf {w}}

\def \vzero{\mathbf {0}}

\def \vA{\mathbf {A}}
\def \vC{\mathbf {C}}

\def \vI{\mathbf {I}}
\def \vM{\mathbf {M}}
\def \vP{\mathbf {P}}
\def \vR{\mathbf {R}}
\def \vL{\mathbf {L}}
\def \vS{\mathbf {S}}
\def \vX{\mathbf {X}}
\def \vU{\mathbf {U}}
\def \vV{\mathbf {V}}
\def \vZ{\mathbf {Z}}

\def \bbR{{\mathbb R}}

\def \cC{{\mathcal C}}
\def \cF{{\mathcal F}}
\def \cN{{\mathcal N}}
\def \cO{{\mathcal O}}
\def \cS{{\mathcal S}}
\def \cE{{\mathcal E}}
\def \cPC{{\mathcal PC}}
\def \Pr{{\rm Pr}}

\newcommand{\jdh}[1]{}
\renewcommand{\jdh}[1]{{\color{red}{#1}}} 

\newtheorem{thmi}{Theorem}[section]

\newtheorem{cori}{Corollary}[section]

\newtheorem{lemmai}{Lemma}[section]

\title{\huge On Convolutional Approximations to
Linear Dimensionality Reduction Operators for Large-Scale Data Processing} 
\author{Swayambhoo Jain~\IEEEmembership{Student Member,~IEEE} and Jarvis Haupt~\IEEEmembership{Member,~IEEE}%
\thanks{The authors are with the Department of Electrical and Computer Engineering at the University of Minnesota -- Twin Cities. Email: \{jainx174, jdhaupt\}@umn.edu. Manuscript submitted February 24, 2015. }%
\thanks{This work was supported in part by DARPA/ONR Grant No. N66001-11-1-4090 and the DARPA Young Faculty Award, Grant No. N66001-14-1-4047.} }

\begin{document}
\maketitle

\begin{abstract}
In this paper, we examine the problem of approximating a general linear dimensionality reduction (LDR) operator, represented as a matrix $\vA \in \bbR^{m \times n}$ with $m<n$, by a partial circulant matrix with rows related by circular shifts. Partial circulant matrices admit fast implementations via Fourier transform methods and subsampling operations; our investigation here is motivated by a desire to leverage these potential computational improvements in large-scale data processing tasks.  We establish a fundamental result, that most large LDR matrices (whose row spaces are uniformly distributed) in fact cannot be well-approximated by partial circulant matrices. Then, we propose a natural generalization of the partial circulant approximation framework that entails approximating the range space of a given LDR operator $\vA\in\bbR^{m\times n}$ over a restricted domain of inputs, using a matrix formed as a product of a partial circulant matrix having $m'>m$ rows and a $m\times m'$ ``post-processing'' matrix.  We introduce a novel algorithmic technique, based on sparse matrix factorization, for identifying the factors comprising such approximations, and provide preliminary evidence to demonstrate the potential of this approach. 
\end{abstract} 


\section{Introduction}

Numerous tasks in signal processing, statistics, and machine learning employ dimensionality reduction methods to facilitate the processing, visualization, and analysis of (ostensibly) high-dimensional data in (more tractable) low-dimensional spaces.  Among the myriad of dimensionality reduction techniques in the literature, \emph{linear dimensionality reduction} (LDR) methods remain among the most popular and widely-used.  

One well-known example is principal component analysis \cite{pearson1901} characterized by a linear dimensionality reduction (LDR) operator designed to maximally preserve the variance of the original data in the projected space, and which has been widely used for data compression, denoising, and as a dimensionality reducing pre-processing step for other analyses (e.g., clustering, classification, etc.) \cite{jolliffe2002principal}.  Other classical data analysis methods that employ specialized LDR operators include linear discriminant analysis (where the operator is designed to preserve separations among original data points belonging to disparate classes), and canonical correlations analysis (where the operator maximizes correlations among  projected data points). See the survey paper \cite{cunningham2014unifying} for additional examples.

In recent years, LDR methods have also been utilized for universal ``precompression'' in high-dimensional inference tasks.  For example, fully random LDR operators are at the essence of the initial investigations into \emph{compressed sensing} (CS) (see, e.g., \cite{Candes:06:Freq, Donoho:06:CS, Candes:06:UES, Haupt:06}); and other, more structured, LDR operators -- both non-adaptive (see, e.g., \cite{elad2007optimized, rudelson2008sparse, ashok2008compressive, duarte2009learning, calderbank2010construction, haupt2010restricted, zelnik2011sensing, ashok2011compressive, carson2012communications, jain2013compressive}) and adaptive (see, e.g., \cite{Ji:08:BCS, Haupt:09:CDS, Haupt:09:CBS, Haupt:11:Chapter, Indyk:11, ashok2011information, Price:12, Iwen:12, Haupt:12:CDS, Davenport:12:CBS, Arias-Castro:13, duarte2013task, Soni:14, Malloy:14, Castro:14}) -- have also been examined recently in the context of CS and sparse inference.   

The computational efficiency of LDR methods is often cited as one of their primary virtues; however, as data of interest become increasingly large-scale (high-dimensional, and numerous), even the relatively low computational complexity associated with LDR methods can become significant.  In this paper, we investigate the utility of employing \emph{partial circulant approximations} to general LDR operators; partial circulant matrices admit fast implementations (via convolution or Fourier transform methods, and subsampling), and their use as surrogates to arbitrary LDR matrices may provide potentially significant computational efficiency improvements in practice.

\subsection{Problem Statement and Our Contributions}

We represent an arbitrary (real) linear dimensionality reduction (LDR) operator as a matrix $\vA \in \bbR^{m \times n}$, with
$m<n$.  Operating with $\vA$ on an arbitrary vector $\vx\in\bbR^n$ generally requires $\cO(mn)$ operations, which can be superlinear in $n$ For even modest values $m$ (e.g., when $m=n^\beta$ for $\beta\in(0,1]$ the complexity is $\cO(n^{1+\beta})$). Here, we seek computationally-efficient approximations of $\vA$ implementable via convolution and downsampling (and, perhaps, modest post-processing).  

We first consider approximating $\vA$ as $\vA\approx\vS\vC$, where $\vC$ is an $n\times n$ circulant matrix and $\vS$ is a matrix with $m$ rows that are a (permuted) subset of $\vI_n$ ($n\times n$ identity).  These \emph{partial circulant} approximations enjoy an implementation complexity of $\cO(n \log n)$, owing to fast Fourier transform methods.  Despite these potential implementation complexity benefits, our first contribution establishes that most LDR matrices, in fact, \emph{cannot} be well-approximated by partial circulant matrices.

We then propose a generalization that uses approximations of the form $\vA\approx\vP\vS\vC$, where $\vC$ and $\vS$ are as above, except that $\vS$ has some $m'\geq m$ rows, and $\vP$ is an $m\times m'$ ``post-processing'' matrix.  Operating with such matrices requires $\cO(mm' + n\log n)$ operations in general, and can be as low as $\cO(n\log n)$, e.g., when $m'=\cO(n^{1/2})$. Within this framework, we also exploit the fact that signals of interest often reside on a \emph{restricted input domain} (e.g., they lie on a union of subspaces, manifold, etc.), so we may restrict our approximation to mimicking the action of $\vA$ on these inputs.  We describe a \emph{data-driven} approach to learning the factors of the approximating matrix in these settings, and provide empirical evidence to demonstrate the viability of this approximation approach. 

\subsection{Connections to Existing Work}
Circulant approximations to square matrices are classical in linear algebra; for example, circulant preconditioners for linear systems were examined in \cite{strang1986proposal, chan1988optimal, tyrtyshnikov1992optimal}.  In a line of work motivated by ``optical information processing,'' several efforts have examined fundamental aspects of approximating square matrices by products of circulant and diagonal matrices \cite{muller1998algorithmic, schmid2000decomposing, huhtanen2007real, huhtanen2007factoring, huhtanen2008approximating}. Here, our focus is on LDR matrices (not square matrices), so results from these works are not directly applicable here. 

Random partial circulant matrices have been studied recently in compressed sensing tasks \cite{romberg2009compressive, haupt2010toeplitz, rauhut2010compressive, rauhut2012restricted, yin2010practical}, and random partial circulant matrices with diagonal pre-processors have been proposed as computationally efficient methods for Johnson-Lindenstrauss (JL) embeddings in \cite{krahmer2011new,  zhang2013new}. The work \cite{yap2013stable} established the viability of using random partial circulant matrices for embedding manifold-structured data.  In contrast to these works, here our aim is to approximate the action of \emph{given} LDR matrix, not necessarily to perform JL embeddings.  
%


\subsection{Outline} 

Following the introduction of a few preliminaries (below), we establish in Sect.~\ref{sec:simple_approach} a fundamental approximation result regarding partial circulant approximations to general LDR matrices.  In Sect.~\ref{sec:generalized_approach} we propose a generalized approach to the partial circulant approximation problem, and provide empirical evidence to demonstrate its efficacy. A brief discussion is provided in Sect.~\ref{sec:conclusions}; proofs appear in the Appendix.

\subsection{Preliminaries}
We introduce some preliminary concepts and notation that will be used throughout the sequel.  First, we define
\begin{equation*}
\vR = \begin{bmatrix}
\vzero_{ (n-1) \times 1 }   &  \vI_{n-1}  \\ 
1   & \vzero_{1 \times (n-1)}
\end{bmatrix},
\end{equation*}
to be the ``right rotation'' matrix, in that post-multiplication of a row vector by $\vR$ implements a circular shift to the right by one position.  Analogously, post-multiplying a row vector by $\vR^T$ implements a circular shift to the left by one position. To simplify notation we write $\vL = \vR^T$; note that  $\vL\vR = \vI_n$.

We represent an  $n \times n$ (real) circulant matrix by 
\begin{equation} \label{eq:circ_form}
\vC = \begin{bmatrix}
		c_1 &	c_2 & \cdots & c_n			\\ 
        		c_{n} & c_1 & \cdots & c_{n-1}	 \\
		 &	   & \ddots & 	\\
		c_2     &	 c_3       & \cdots & c_1 
	\end{bmatrix} = 
	 \begin{bmatrix}
		\vc^T \\
		 \vc^T \vR \\
		 \vdots \\ 
		 \vc^T \vR^{n-1} 
	\end{bmatrix}, 
\end{equation} 
where $\vc = [c_1 \ \cdots \ c_n]^T \in \bbR^n$.  We let $\cC_n$ denote the set of all (real) $n$-dimensional circulant matrices of the form \eqref{eq:circ_form}.
For $m < n$, the set of all $m \times n$ real \emph{partial} circulant matrices is
\begin{equation*}\label{eq:circ_set}
\cPC_{m,n} = \left\{ \vS \vC \in \bbR^{m\times n} \ | \  \vS \in \cS_m, \vC \in \cC_n    \right\},
\end{equation*}
where $\cS_m$ is the set of $m \times n $ row sampling matrices whose rows comprise $m$ different canonical basis vectors of $\bbR^n$. 

For a matrix $\vA\in\bbR^{m\times n}$, we denote its $m$ individual rows by $\vA_{i,:}$ for $i=1,2,\dots,m$ and its $n$ columns by $\vA_{:,j}\in\bbR^m$ for $j=1,\dots,n$.  We write the squared Frobenius norm of $\vA$ as $\|\vA\|_F^2 = \sum_{i,j} |A_{i,j}|^2$, and its $1,2$ norm by $\|\vA\|_{1,2} = \sum_{j=1}^n \|\vA_{:,j}\|_2$, where $\|\vA_{:,j}\|_2$ denotes the Euclidean norm of $\vA_{:,j}$.  

\section{A Fundamental (Negative) Approximation Result for Partial Circulant Matrices}\label{sec:simple_approach}
As above, we let $\vA \in \bbR^{m \times n}$ denote the LDR matrix we aim to approximate using a partial circulant matrix of the same dimensions.  
Here, we consider a tractable (and somewhat natural) choice of the approximation error metric, and seek to find the matrix $\vZ \in \cPC_{m,n}$ closest to $\vA$ in the Frobinius sense. 
In this setting, the minimum approximation error is given by
\begin{equation}\label{eqn:circerr}
\cE_{\cPC_{m,n}}(\vA) =   \min_{ \vZ \in \cPC_{m,n}} \ \|\vA - \vZ\|_F^2.
\end{equation}
Finding the minimizer of \eqref{eqn:circerr} is a non-convex optimization problem, owing to the product nature of elements of $\cPC_{m,n}$.  Nevertheless, we obtain a precise characterization of the minimum achievable approximation error for a given $\vA$ via the following lemma (whose proof appears in the appendix).

\begin{lemmai}\label{lem:min_fro_error} 
\emph{For $\vA \in \bbR^{m \times n}$, we have 
\begin{equation}\label{eqn:rubikerr}
\cE_{\cPC_{m,n}}(\vA) = \|\vA\|_F^2 - R^2(\vA), 
\end{equation}
where
\begin{equation} \label{eqn:rubik}
R(\vA) = \max_{\vf \in \cF} \frac{ \| \sum_{i=1}^m  \vA_{i,:} \vL^{f_i}   \|_2 }{\sqrt m},
\end{equation}
and 
$\cF = \left\{ \vf = [f_1 ... f_m] \in \{0,...,n-1\}^m\Big |  f_i \ne f_j \ \forall i \ne j  \right\}$.}
\end{lemmai}

We call the term $R(\vA)$ in \eqref{eqn:rubik} the \emph{Rubik's Score} of the matrix $\vA$,  inspired by the fact that $R(\vA)$ is maximized when the circular shifts of the rows $\{\vA_{i,:}\}$ are ``maximally aligned''  (indeed, the numerator of \eqref{eqn:rubik} is a sum of rotated rows of $\vA$).   

Evidently, $ 0 \le R(\vA) \le \|\vA\|_F$, with $R(\vA) = \| \vA  \|_F$ for all $\vA \in \cPC_{m,n}$, but beyond these ``boundary'' cases, Lemma~\ref{lem:min_fro_error} yields limited interpretational insight into the achievable error for any particular $\vA$.  We gain additional insight here using a probabilistic technique -- instead of quantifying the approximation error for a fixed $\vA$, we consider instead matrices $\vA$ whose row spaces are distributed uniformly at random on ${\rm Gr}(m,n)$, the Grassmannian manifold of $m$-dimensional linear subspaces of $\bbR^n$.  Using this model, we  can quantify the \emph{proportion} of matrices so drawn whose optimal partial circulant approximation error is at most a fixed fraction (say $\delta$) of their squared Frobenius norm.

To this end, we exploit the fact that matrices whose row-spaces are uniformly distributed on ${\rm Gr}(m,n)$ may be modeled as matrices having iid zero-mean Gaussian elements. With this, we establish the following theorem (proved in the appendix).
\begin{thmi}\label{thrm:Prob_bound_theorem} 
\emph{For $2 \le m \le n$, let $\vA \in \bbR^{m \times n}$ have iid $\cN(0,1)$ entries. Then for $\delta\in[0,0.125)$,
and $n$ is sufficiently large,
there exists a positive constant $c(\delta)$ such that
\begin{equation*}
\Pr( \cE_{\cPC_{m,n}}(\vA) \le \delta \| \vA  \|_F^2 ) = \cO  (e^{-c(\delta) \cdot m n}  ). 
\end{equation*}}
\end{thmi}

Simply put, the content of Theorem~\ref{thrm:Prob_bound_theorem} is that the proportion of large matrices (with uniformly distributed row spaces) that can be approximated to high accuracy by partial circulant matrices is exponentially small in the product of the matrix dimensions.  In the next section we propose a more general framework designed to facilitate accurate partial circulant approximations in a number of practical applications.

\section{A Generalized ``Data-Driven'' Partial Circulant Approximation Approach}\label{sec:generalized_approach}

\begin{algorithm}[t!]
	\caption{``Data-Driven'' Partial Circulant Approximation}
	\label{ag}
	\begin{algorithmic} 
		\STATE \hspace{-1.4em} \textbf{Inputs:} $\mbox{LDR matrix } \vA\in\bbR^{m\times n}$, $\mbox{parameters } \lambda, \mu, \epsilon > 0$,
		\STATE \hspace{2.1em} $\mbox{Matrix of ``representative'' data } \vX\in\bbR^{n\times p}$,
		\STATE \hspace{-1.4em} \textbf{Initialize:} $\vM^{(0)} = \vU \Sigma$ (from the SVD $\vA \vX= \vU \Sigma \vV^T$)
		\STATE \hspace{3.0em} $\textrm{obj}^{(0)} = \|\vA\vX\|_F^2$
		\REPEAT 
		\STATE \hspace{-0.9em} $\vC^{(t)} = \arg \min_{\vC \in \mathcal{C}_n} \|\vA\vX - \vM^{(t-1)} \vC \vX\|_F^2 + \mu \| \vC \|_F^2 $
		\STATE \hspace{-0.6em}$\vM^{(t)} = \arg \min_{\vM \in \bbR^{m \times n} } \|\vA\vX - \vM\vC^{(t)}\vX\|_F^2 + \lambda \| \vM  \|_{2,1}$
		\STATE \hspace{-0.6em}$\textrm{obj}^{(t)} = \|\vA\vX - \vM^{(t)}\vC^{(t)}\vX\|_F^2 + \mu \| \vC^{(t)} \|_F^2 + \lambda \| \vM^{(t)}  \|_{2,1}$
		\UNTIL  $ \textrm{obj}^{(t)} - \textrm{obj}^{(t-1)} \le \epsilon \cdot \textrm{obj}^{(t-1)}$
		\STATE \hspace{-1.4em} \textbf{Output:} $\vM^* = \vM^{(t)}, \vC^* = \vC^{(t)} $\\ 
	\end{algorithmic}
\end{algorithm}

That the conclusion of Theorem~\ref{thrm:Prob_bound_theorem} be pessimistic is, perhaps, intuitive, since $\vA\in\bbR^{m\times n}$ has $\cO(mn)$ ``degrees of freedom,'' while any $\vC\in\cPC_{m,n}$ has only $\cO(n + m)$.   These structural limitations are, of course, fundamental. We gain additional leverage here by exploiting a more general approximation model, and exploiting additional latent ``signal space characteristics'' that arise in many practical applications.  

Formally, we consider approximations of $\vA$ of the form $\vA \approx \vP \vS \vC$ where $\vC\in\cC_n$, $\vS$ is an $m'\times n$ matrix comprising a permuted subset of rows of identity, and $\vP \in \bbR^{m \times m'}$ is a ``post-processing'' matrix.  This extension allows approximations of $\vA$ whose rows are generally \emph{linear combinations} of the rows of a partial circulant matrix rather than being rows of a partial circulant matrix themselves, facilitating accurate approximation of any matrix whose row space is \emph{spanned by} vectors related by circular shifts.

Further, we exploit the fact that in many applications where LDR methods are deployed (e.g, PCA, LDA, Compressive Sensing, etc.), the high-dimensional data to be processed are not arbitrary, but instead lie in some \emph{restricted input domain} (e.g., they lie in a low-dimensional subspace, a union of low-dimensional subspaces, distinct clusters, etc.).  In these cases, our approximation task reduces to a simpler task of mimicking the action of $\vA$ on these restricted inputs. 

Suppose $\vX\in\bbR^{n\times p}$ is a matrix whose columns are ``representative'' of the restricted input domain for the problem of interest. Then, minimizing an objective of the form $\|\vA\vX - \vP\vS\vC\vX\|_F^2$  essentially ensures the approximation be accurate in an (empirical) \emph{average} sense. We propose an algorithmic approach based on alternating minimization and regularized matrix factorization for identifying the factors comprising this more general approximation; the method is summarized as Algorithm~\ref{ag}.  Note that in our approach, we effectively combine the action of the sampling and post processing matrices; we let $\vP\vS\triangleq \vM$, and seek \emph{column sparsity} in $\vM$ using the $\| \vM  \|_{2,1}$ regularization term, which penalizes the number of nonzero columns of $\vM$ (the resulting number $m'$ of nonzero columns depends on the regularization parameter $\lambda>0$). We also include a (simple) regularization term for $\vC\in\bbR^{n\times n}$ to fix scaling ambiguities; leaving $\vC$ unconstrained would allow elements of $\vM$ to become arbitrarily small, and would circumvent the effect of the column-sparsity regularization.  Both of these sub-problems are convex, and can be solved using off-the-shelf software (e.g., SLEP \cite{Liu:2009:SLEP:manual}).

\begin{figure}[t]
	\hspace{-0.6em}\begin{tabular}{ccc}
		\includegraphics[width=0.3 \linewidth]{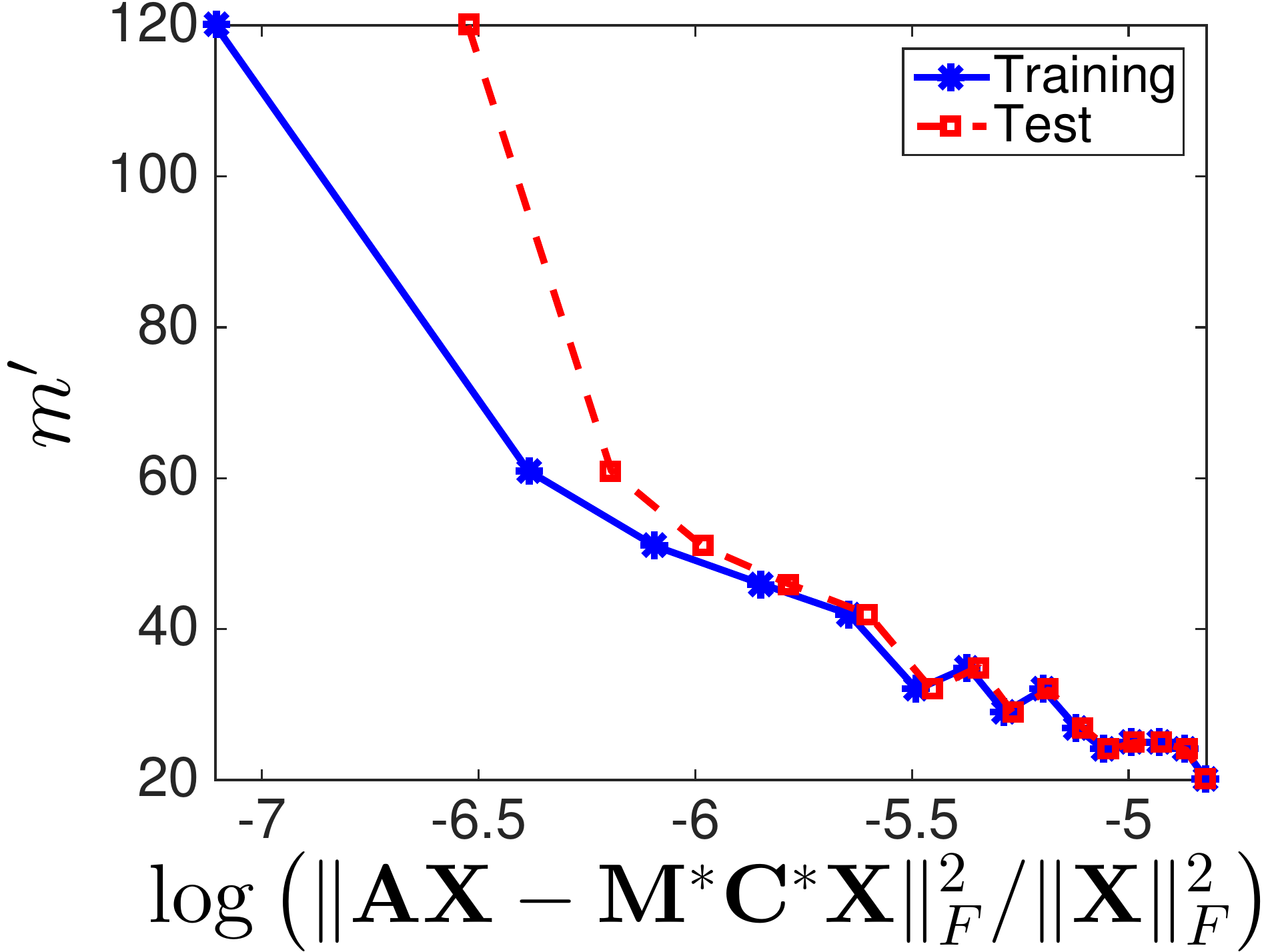} &
		\includegraphics[width=0.3 \linewidth]{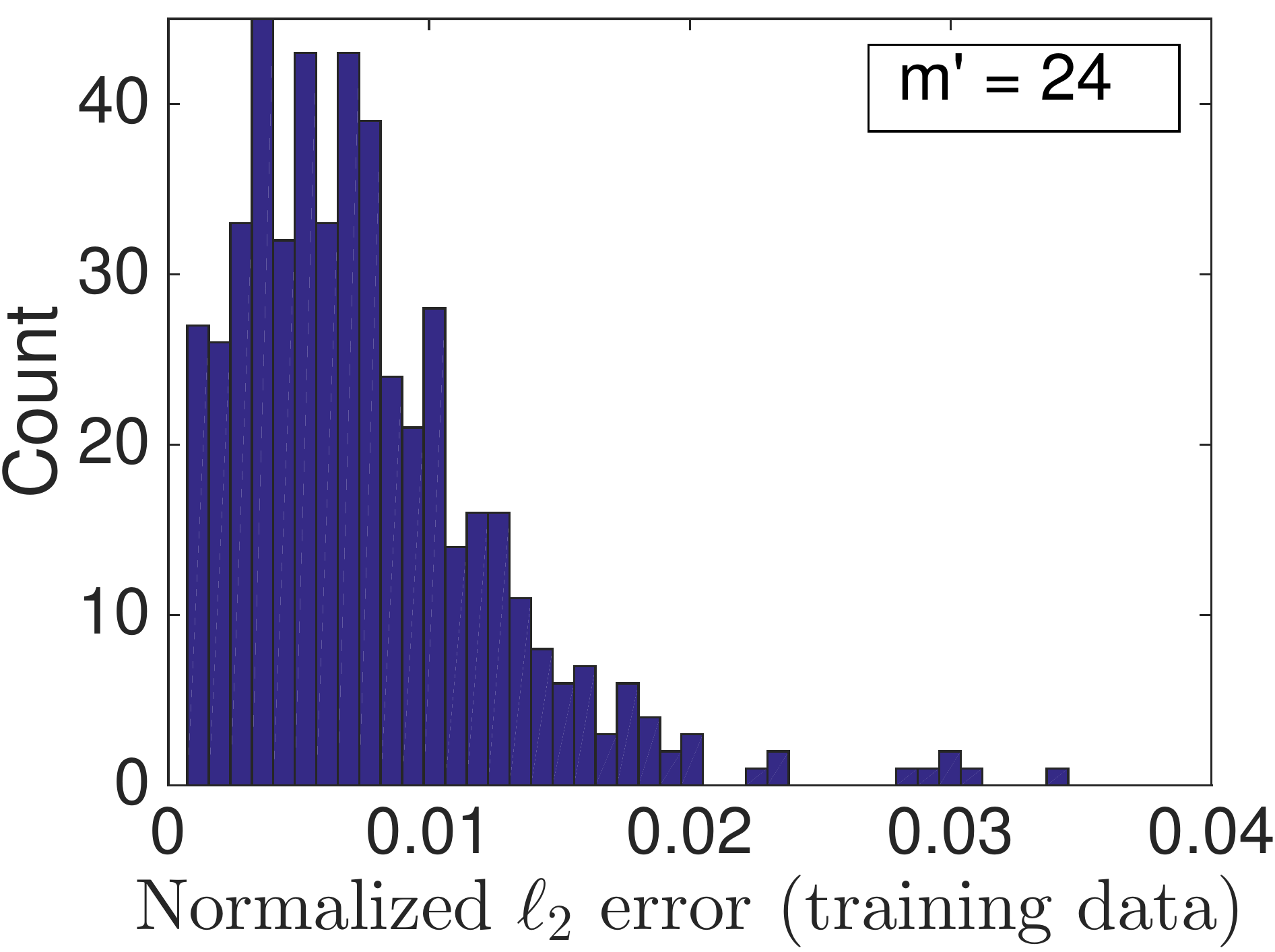} &
		\includegraphics[width=0.3 \linewidth]{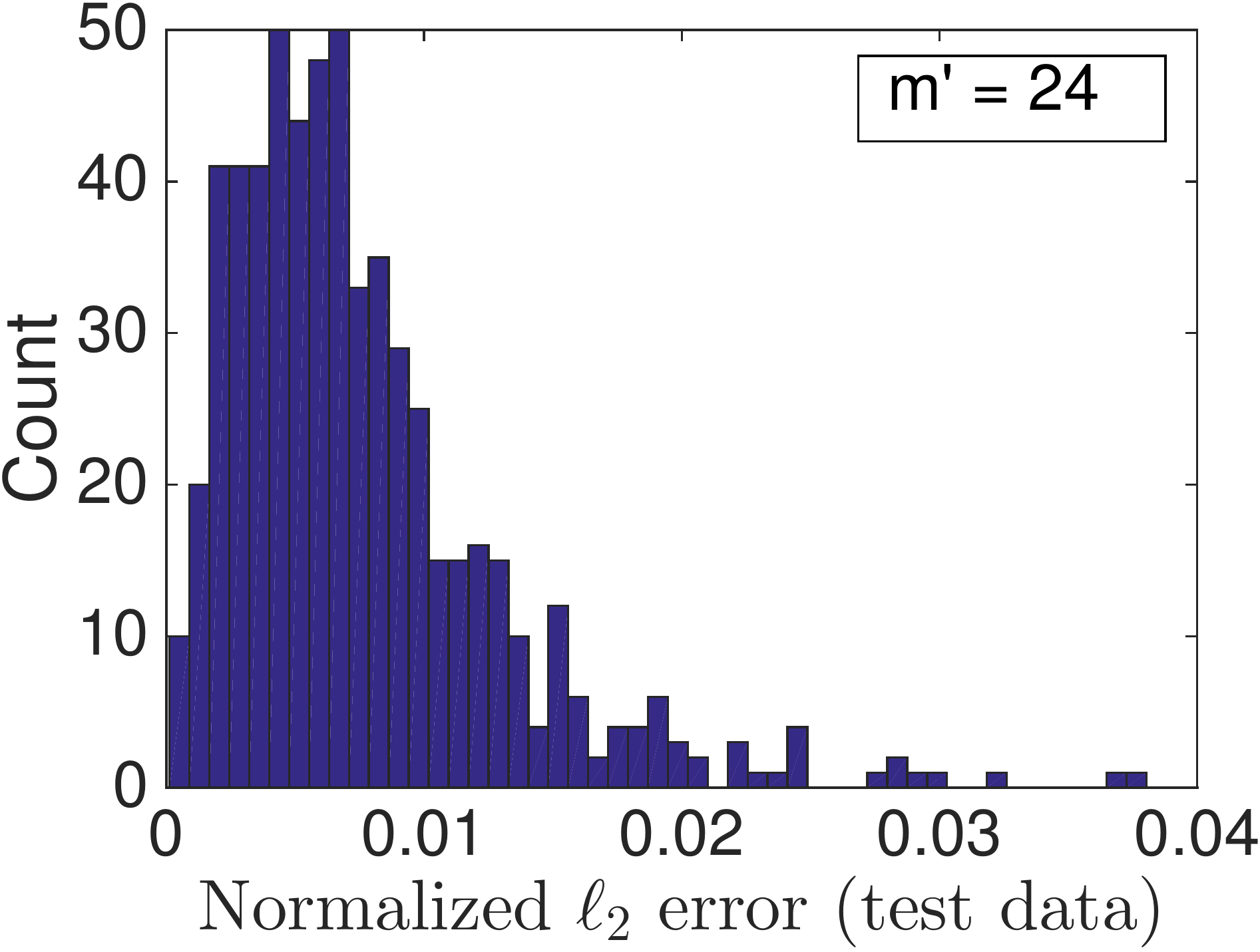} \\
		\includegraphics[width=0.3 \linewidth]{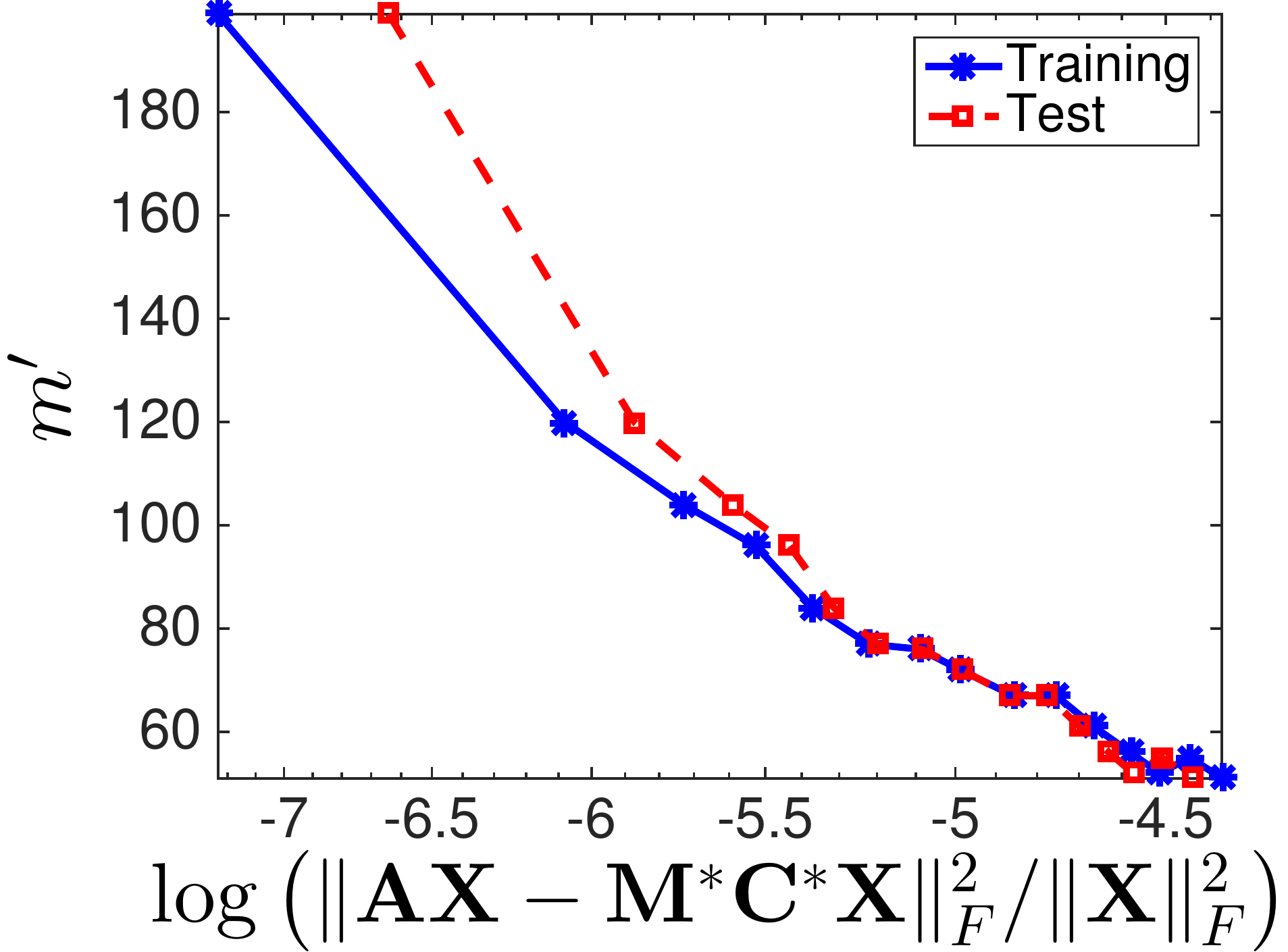} &
		\includegraphics[width=0.3 \linewidth]{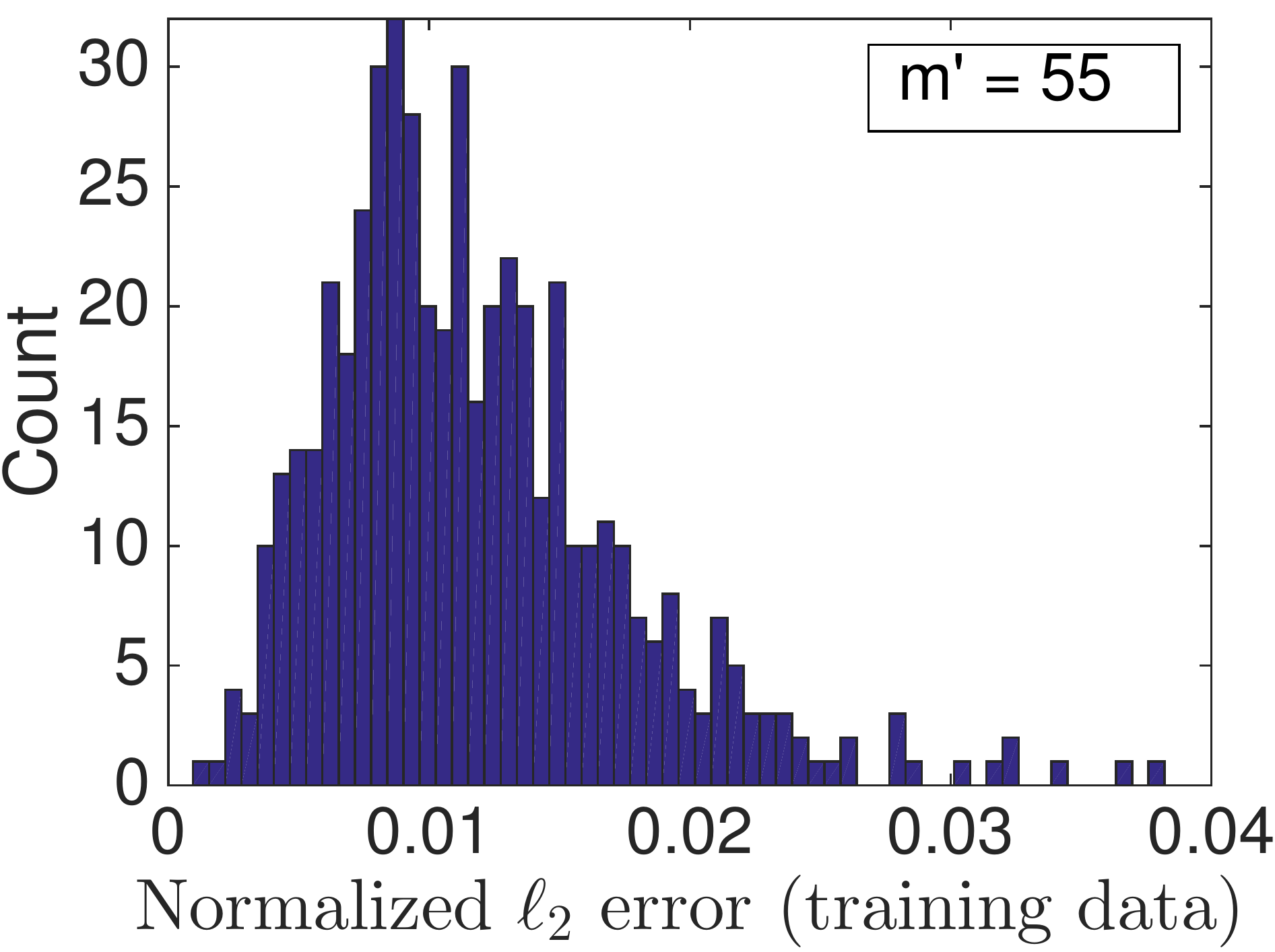} &
		\includegraphics[width=0.3 \linewidth]{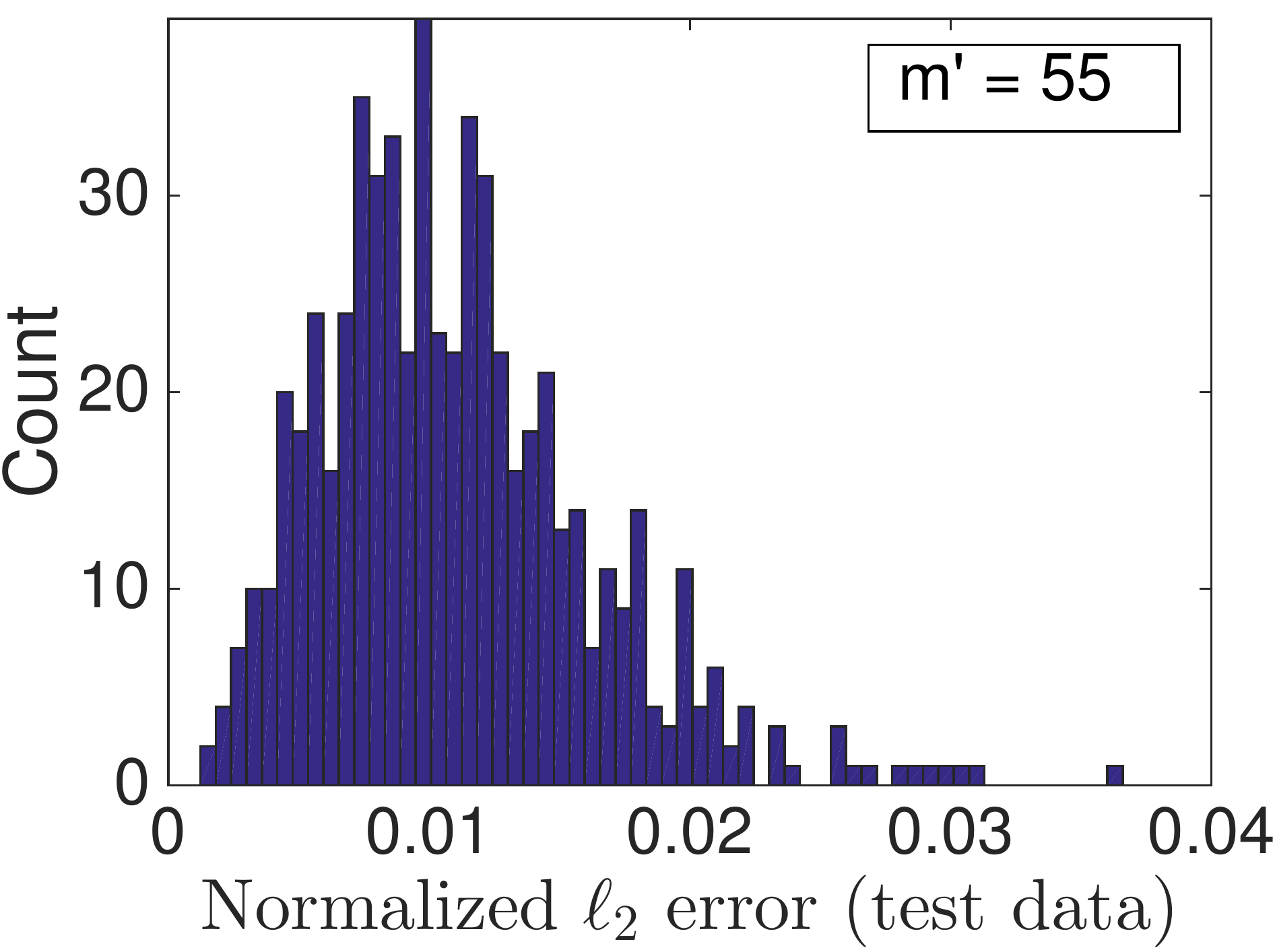} \\
	\end{tabular}
	\caption{Data-driven approximation of matrices comprised of top principal components of a training set of `$2$' digits from the USPS dataset.  The panels are (from left to right) average approximation error vs. $m'$, histogram of normalized training errors, and histogram of normalized test data errors.  The top and bottom rows correspond to matrices $\vA$  whose rows are the top $5$ and top $10$ principal component direction vectors, respectively.}
	\label{fig:res1}
\end{figure}

To evaluate the performance of this approach, we chose $\vX$ to be a matrix of training data whose columns are $500$ vectorized $16\times 16$ images of the digit `$2$' from the USPS handwritten digit dataset (available at \url{http://www.cs.nyu.edu/~roweis/data.html}), and held out the remaining images of the digit `$2$' to form a ``test'' data set. We select the rows of the matrix $\vA$ to be a subset of the top principal component vectors of the training set $\vX$, and we consider two settings corresponding to $\vA$ having $5$ rows and $10$ rows. Then, we applied the procedure of Alg.~\ref{ag}, with $\mu=0.1$ and varying values of $\lambda$ to obtain different factorizations $\vM^*$ and $\vC^*$. For each, we quantified the normalized error on the training set (on a logarithmic scale) vs. the column sparsity of $\vM^*$.  For a convenient choice of $\lambda$, we also computed histograms of the normalized approximation errors ($\|\vA\vx - \vM\vC\vx\|_2^2/\|\vx\|_2^2$) for each point in the testing and training data sets.  The results are provided in Fig.~\ref{fig:res1}, and illustrate that accurate approximations can indeed be obtained for reasonable values of $m'$ for each case, and that the approximations generalize well to the test data, in that the approximation error histogram is concentrated around small values for each ($\preceq 2\%$ error for most points).

\section{Discussion and Conclusions}\label{sec:conclusions}

In this work we established a fundamental result regarding partial circulant approximations of general matrices, and proposed and demonstrated a general approach for approximating the action of general linear dimensionality reduction (LDR) operators over restricted input domains.   The convolutional nature of the approximations we consider here, in addition to their computational expediency, makes them viable for implementation using LTI systems, or for specialized applications (e.g., RADAR) where convolutional models arise naturally. 

A more general investigation motivated by computational complexity considerations would also include other specialized matrix structures with low implementation complexities (e.g., sparse matrices or ``fast'' or sparse Johnson-Lindenstrauss embeddings \cite{Ailon:06, Dasgupta:10}).  A thorough, unifying analysis of such ``low-complexity'' approximations is the topic of our ongoing work, and will be reported in a future publication.

\appendix
\subsection{Proof of Lemma~\ref{lem:min_fro_error}} We first write $\cE_{\cPC_{m,n}}(\vA)$ in \eqref{eqn:circerr} equivalently as 
\begin{align}\label{eq:opt_prob_orig}
\cE_{\cPC_{m,n}}(\vA) = \min_{\vS \in \cS_m} \min_{ \vC \in \cC_n } \ \|\vA - \vS\vC\|_F^2.
\end{align}
Now, a key insight is that each choice of $\vS\in\cS_m$ corresponds to a vector $\vf$ from the set $\cF$ defined in Lemma~\ref{lem:min_fro_error}, as including the $i$-th row of $\vI_n$ in $\vS$ corresponds to selecting the $i$-th row of $\vC$, which is given from \eqref{eq:circ_form} by $\vc^T \vR^{i-1}$.  Using this we may parameterize the choice of $\vS$ in terms of $\vf$, and rewrite the objective function in \eqref{eq:opt_prob_orig} as 
\begin{eqnarray*}
\lefteqn{ \|\vA - \vS\vC\|_F^2 = \sum_{i=1}^{m} \left\| \vA_{i,:} - \vc^T \vR^{f_i}   \right\|_2^2}&&  \\ 
&=&  \sum_{i=1}^{m} \| \vA_{i,:}\|_2^2 + \vc^T \vR^{f_i} \vL^{f_i}  \vc - 2 \vA_{i,:} \vL^{f_i} \vc.
\end{eqnarray*}
Thus, since $ \vR^{f_i} \vL^{f_i} = \vI_n$, we have  
\begin{equation}\label{eqn:err2}
\cE_{\cPC_{m,n}}(\vA)  = \min_{\vf \in \cF} \min_{ \vc \in \bbR^n } \ \| \vA \|_F^2 + m\vc^T\vc - 2\left(\sum_{i=1}^{m}  \vA_{i,:} \vL^{f_i}  \right) \vc.
\end{equation}

We first minimize the objective with respect to $\vc$, keeping $\vf$ fixed to any arbitrary value in $\cF$.  This is an unconstrained strictly convex quadratic problem whose minima can be obtained by equating the gradient (with respect to $\vc$) to zero.  We identify the optimal $\vc$ to be $ \vc^* = \frac{1}{m} \sum_{i=1}^{m}  \vR^{f_i} (\vA_{i,:})^T$.  Substituting $\vc^*$ into \eqref{eqn:err2}, and simplifying, we obtain
\begin{equation*}
\cE_{\cPC_{m,n}}(\vA) = \min_{\vf \in \cF} \|\vA\|_F^2 - \frac{1}{m}  \left\| \sum_{i=1}^{m}  \vA_{i,:} \vL^{f_i} \right\|_2^2,
\end{equation*}
which gives \eqref{eqn:rubikerr}, using the definition \eqref{eqn:rubik} of the \emph{Rubik's Score}.


\subsection{Proof of Theorem \ref{thrm:Prob_bound_theorem}}


Note that $|\cF| = n!/(n-m)!$ (since it is just the number of ways of choosing $m$ elements out of $n$ without replacement), and we have (trivially) that $n!/(n-m)! < n^m$, so 
%
\begin{eqnarray*}
\lefteqn{\Pr( \cE_{\cPC_{m,n}}(\vA) \le \delta \|  \vA \|_F^2 )}&&\\
&=& \Pr\left( (1-\delta) \|  \vA \|_F^2  \le  \sup_{f \in \cF} \frac{ \| \sum_{i=1}^m \vA_{i,:} \vL^{f_i}   \|_2^2 }{ m} \right)  \\
&=& \Pr\left(  \bigcup_{f \in \cF} \left[ (1-\delta)   \| \vA  \|_F^2  \le   \frac{ \| \sum_{i=1}^m \vA_{i,:} \vL^{f_i}  \|_2^2 }{ m}  \right] \right) \\ 
&\le& n^m \ \Pr\left(  (1-\delta)\| \vA  \|_F^2  \le   \frac{ \| \sum_{i=1}^m \vA_{i,:} \vL^{f_i}   \|_2^2 }{ m}   \right), 
\end{eqnarray*}
where the last step follows from union bounding.

Now, we can simplify further by introducing the shorthand notation $\va = [\vA_{i,:} \ \vA_{2,:} \ \dots \ \vA_{m,:}]^T\in\bbR^{mn}$ and $\tilde{\vR}_{\vf} = [\vR^{f_1} \ \vR^{f_2} \ \dots \vR^{f_m}]\in\bbR^{n\times mn}$, so that $\|\vA\|_F^2 = \va^T\va$ and $\|\sum_{i=1}^m \vA_{i,:} \vL^{f_i}\|_2^2/m = \va^T (\tilde{\vR}_{\vf}^T \tilde{\vR}_{\vf}/m) \va$. It is easy to check that $(\tilde{\vR}_{\vf}^T  \tilde{\vR}_{\vf} /m) \tilde{\vR}_{\vf}^T  = \tilde{\vR}_{\vf}^T$, which implies that the columns of $\tilde{\vR}_{\vf}^T$ are eigenvectors of the matrix corresponding to eigenvalue $1$. Further, since $(\tilde{\vR}_{\vf}^T  \tilde{\vR}_f /m)$ is symmetric, it admits an eigendecomposition $(\tilde{\vR}_{\vf}^T  \tilde{\vR}_{\vf}/m)  = \vU_{\vf} \Sigma_{\vf} \vU_{\vf}^T$ with $\vU_{\vf}$ orthonormal and $\Sigma_{\vf}$ diagonal, and since it is rank $n$, $\Sigma_{\vf}$ has exactly $n$ entries being $1$ (and the rest $0$).  

Incorporating this insight into the probabilistic analysis above, we obtain that
\begin{eqnarray*}
\lefteqn{\Pr\left(  (1-\delta)\| \vA  \|_F^2  \le   \frac{ \| \sum_{i=1}^m \vA_{i,:} \vL^{f_i}   \|_2^2 }{ m}   \right) }&&\\
&=&  \Pr\left(\va^T \vU_{\vf} \ ((1-\delta) \vI_{mn} - \Sigma_{\vf} )) \  \vU_{\vf}^T \va \leq 0\right)\\
&=& \Pr\left( \tilde{\va}^T  \ \left( \left (1-\delta \right) \vI_{mn} - \Sigma_{\vf}  \right) \ \tilde{\va}  \le 0   \right),
\end{eqnarray*}
where the components of $\tilde{\va}=\vU_{\vf} \va$ are iid $\cN(0,1)$ due to the unitary invariance of the Gaussian distribution.  Thus, with a slight overloading of notation, we may write that
\begin{eqnarray}\label{eqn:ssproj} 
\nonumber \lefteqn{\Pr( \cE_{\cPC_{m,n}}(\vA) \le \delta \|  \vA \|_F^2 )}&&\\
&\leq&n^m \ \Pr\left(  \sum_{i=2}^m \| \vA_{i,:} \|_2^2  \le  \delta \sum_{i=1}^m \| \vA_{i,:} \|_2^2 \right).
\end{eqnarray}
%
%
%
%

At this point, we note that vectorizing (row-wise) a random matrix $\vA\in\bbR^{m\times n}$ having iid zero-mean Gaussian elements yields an $mn$-dimensional vector whose direction is selected uniformly at random from ${\rm Gr}(1,mn)$.  Thus, $\sum_{i=1}^m \| \vA_{i,:} \|_2^2$ quantifies the length of the vector, while $\sum_{i=2}^m \| \vA_{i,:} \|_2^2$ describes the energy retained after projecting the vector onto the fixed $n(m-1)$-dimensional subspace spanned by the last $n(m-1)$ coordinates. It follows that the probability on the right-hand side of \eqref{eqn:ssproj} may be interpreted in terms of the energy retained after projecting a \emph{fixed} $mn$-dimensional unit-normed vector onto a subspace selected uniformly at random from ${\rm Gr}(n,mn)$.  To quantify this, we use the following.
\begin{lemmai}[Adapted from Thm. 2.14 of \cite{hopcroftfoundations}]  \label{lem:chi_concentration}
Let $\vv$ be a fixed unit-length vector in $\bbR^d$, $W$ a randomly oriented $k$-dimensional subspace, and $\vw$ the projection of $\vv$ onto $W$. For $0\leq \varepsilon \leq 1$, 
\begin{equation*}
\Pr\left( \|\vw\|_2 \le (1-\varepsilon)\sqrt{k/d}   \right) \le  3e^{- k\varepsilon^2/64}.
\end{equation*}
\end{lemmai}
Using this result (with $k=n(m-1)$ and $d=mn$) we obtain that for $\delta < 1-1/m$,
\begin{eqnarray*}
\lefteqn{\Pr( \cE_{\cPC_{m,n}}(\vA) \le \delta \|  \vA \|_F^2 ) \leq n^m \Pr\left(  \frac{\sum_{i=2}^m \| \vA_{i,:} \|_2^2}{\sum_{i=1}^m \| \vA_{i,:} \|_2^2}  \le  \delta  \right)}\hspace{9em}&&\\ 
&\leq& 3 n^m e^{- \frac{n(m-1)}{64} \left( 1- \sqrt{\frac{\delta}{ 1 - \frac{1}{m} }} \right)^2 }\\
&\leq& 3 \ e^{m\log n - \frac{n(m-1)}{64} \left( 1- 2\sqrt{\frac{\delta}{ 1 - \frac{1}{m} }} \right) }.
\end{eqnarray*}
Now, it is straightforward to verify that for $\delta < 1/8 = 0.125$ and $n$ sufficiently large, so that 
$n/\log n > 128/(1-2\sqrt{2\delta})$,
there exists a positive constant
\begin{equation*}
c(\delta)< \frac{1}{128} \left(1-2\sqrt{2\delta}\right) - \frac{\log n}{n}
\end{equation*}
for which
\begin{equation*}
m\log n - \frac{n(m-1)}{64} \left( 1- 2\sqrt{\frac{\delta}{ 1 - \frac{1}{m} }} \right) \leq  -c(\delta) mn.
\end{equation*}
In this case, it follows that
\begin{equation*}
\Pr( \cE_{\cPC_{m,n}}(\vA) \le \delta \|  \vA \|_F^2 ) \leq 3 e^{-c(\delta) \cdot m n}.
\end{equation*}

\bibliographystyle{IEEEbib}
\bibliography{CAM_Conf_v1}
\end{document}